\documentclass{article}
\usepackage{arxiv}
\usepackage{graphicx}
\usepackage{algpseudocode}
\usepackage{amsmath}
\usepackage{cite}
\usepackage{float}
\usepackage{amsmath,amssymb,amsfonts}
 
\usepackage{subcaption}
\usepackage[ruled,lined,linesnumbered,commentsnumbered,boxed]{algorithm2e}
\usepackage{graphicx}
\usepackage{textcomp}
\usepackage{xcolor}
\usepackage{mathtools}
\usepackage{multirow}
\usepackage{hyperref}
\usepackage{array}
\usepackage{comment}
\usepackage{physics}

\usepackage{arxiv}
\usepackage{fullpage}
\usepackage[utf8]{inputenc}
\usepackage[T1]{fontenc}
\usepackage{hyperref}
\usepackage{url}
\usepackage{amsfonts}
\usepackage{physics}
\usepackage{graphicx}
\usepackage{caption}
\usepackage{float}
\usepackage{subcaption}
\usepackage{mathtools}
\usepackage{comment}
\usepackage[title]{appendix}

\usepackage{amsmath,amsfonts,amssymb,amsthm}
\usepackage{nicefrac}
\usepackage{booktabs}
\theoremstyle{plain}

\theoremstyle{definition}

\theoremstyle{remark}

\newtheorem*{theorem*}{Theorem}
\newtheorem*{lem*}{Lemma}
\newtheorem*{thm*}{Theorem}
\newtheorem*{defn*}{Definition}

\hypersetup{
    colorlinks=true,
    linkcolor=blue,
    filecolor=magenta,      
    urlcolor=cyan,
}

\title{Hamiltonian Monte Carlo Particle Swarm Optimizer}

\author{
  Omatharv Bharat Vaidya \\
  Department of CSIS \& Mathematics \\
  Birla Institute of Technology and Science\\
  Goa, India \\
  \texttt{f20180354@goa.bits-pilani.ac.in} \\
   \And
  Rithvik Terence DSouza \\
  Department of CSIS \& Physics \\
  Birla Institute of Technology and Science\\
  Goa, India \\
  \texttt{f20170200@goa.bits-pilani.ac.in} \\
   \And
  Snehanshu Saha \\
  Department of CSIS and APPCAIR \\
  Birla Institute of Technology and Science\\
  Goa, India \\
  \texttt{snehanshus@goa.bits-pilani.ac.in} \\
  \And
    Soma Dhavala \\
  Founder\\
  MLSquare\\
  Bengaluru, India \\
  \texttt{soma@mlsquare.org} \\
    \And
  Swagatam Das \\
  Electronics and Communication Sciences Unit \\
  Indian Statistical Institute\\
  Kolkata, India \\
  \texttt{swagatam.das@isical.ac.in}
}

\begin{document}
\date{}
\maketitle

\begin{abstract}
We introduce the Hamiltonian Monte Carlo Particle Swarm Optimizer (HMC-PSO), an optimization algorithm that reaps the benefits of both Exponentially Averaged Momentum PSO and HMC sampling. The coupling of the position and velocity of each particle with Hamiltonian dynamics in the simulation allows for extensive freedom for exploration and exploitation of the search space. It also provides an excellent technique to explore highly non-convex functions while ensuring efficient sampling. We extend the method to approximate error gradients in closed form for Deep Neural Network (DNN) settings. We discuss possible methods of coupling and compare its performance to that of state-of-the-art optimizers on the Golomb's Ruler problem and Classification tasks.

\end{abstract}
\section{Introduction}
\label{sec:Introduction}

Particle Swarm Optimization (PSO) techniques have been shown in the past to be analogous to Stochastic Gradient Descent (SGD) both qualitatively and quantitatively \cite{adaswarm}. This has opened up the possibility of training deep neural networks (DNNs) with a diverse class of loss functions, since PSO techniques do not rely on gradients in the way that SGD-based optimizers do. Vanilla-PSO itself has evolved since its proposal in 1995. For instance, Exponentially Averaged Momentum PSO (EM-PSO) is a PSO variant momentum leads to faster convergence. Further, equivalence between EM-PSO and SGD can be used to approximate gradients by harvesting EM-PSO particle position information and fuse it with SGD.  Such an approach was able to outperform the Adam optimizer on a variety of datasets and DNNs \cite{adaswarm}. In the same spirit of coupling different families of optimization techniques,  it has been recognized recently that learning rates can be transferred across training regimes. This grating, facilitates decomposing and recomposing optimizers, and better hyperparameter tuning in DNN optimization tasks\cite{grafting}. However, one of the potential drawbacks of EM-PSO, in general, is that it does not explore the search space well enough and it may settle in the local minima because of its first-order nature. In order to remedy this problem, we propose coupling PSO with a Markov Chain Monte Carlo (MCMC) sampling technique such as Hamiltonian Monte Carlo (HMC) \cite{hmc}. The hope is that, this grafted opimizer-cum-sampler can better navigate the explore-exploit tradeoff that is prevalent in DNN training. Recall that HMC is used for sampling from probability distributions by simulating particles in a potential, subjected to Hamiltonian dynamics. Coupled with a conditional proposal acceptance step, this method produces a stationary distribution that approximates the original probability distribution fairly well. It explores the parameter space substantially and ensures that a global solution is found for the optimization problem. We discuss few possible ways of coupling HMC with PSO and report the findings on various multi-objective optimization as well as deep learning problems. We have observed that the modified optimizer makes the loss drop faster and leads to better testing accuracy on DNNs.

We term our proposed algorithm Hamiltonian Monte Carlo Particle Swarm Optimization (HMC-PSO). We demonstrate its effectiveness in solving a non-traditional yet important problem like the Golomb's ruler. Also, we compare the DNN version of HMC-PSO with other state-of-the-art (SOTA) optimizers like Adam, Adadelta, RMSprop, Adagrad, etc., on some important classification problems and show SOTA results. We begin with the background and motivation for the proposed HMC-PSO in Section $2$. Required theoretical context, setup, and proposal for HMC-PSO are presented in Section $3$. We showcase the experimental results and comparisons with current SOTA optimizers in Section $4$. Finally, Section $5$ states the conclusion.

\section{Motivation and Background}
\label{sec:Background}

The sources of inspiration to understand and improve optimization techniques in general, and Particle Swarm Optimization (PSO) in particular, come from several places. Said et al. \cite{mikkiphysics} postulated that swarms behave similarly to classical and quantum particles. In fact, their analogy is so striking that one may tend to think that the social and individual intelligence components in PSO are, after all, nice useful metaphors and that there is a neat underlying dynamical system at play. This dynamical system perspective was indeed useful in unifying two almost parallel streams, namely, optimization and Markov Chain Monte Carlo sampling. In a seminal paper, Wellington and Teh \cite{wellingdynamics}, showed that a stochastic gradient descent (SGD) optimization technique could be turned into a sampling technique by just adding noise, governed by Langevin dynamics. Recently, Soma and Sato \cite{satoito} provided further insights into this connection based on an underlying dynamical system governed by stochastic differential equations (SDE). These new results make a strong case for connections between optimization and sampling based on Stochastic Approximation and Finite Differences and made us wonder: Is there a larger, more general template of which the aforementioned approaches are special cases? Can we develop a unified dynamical system that can be made to operate either sampling mode (in the MCMC sense) or search model (in the SGD sense)? Precisely, can we
\begin{itemize}
\item Develop a unifying coupled-dynamical system that can couple existing paradigms such as an SGD and HMC, SGD and PSO, HMC and PSO, etc.
\item Using the coupled-dynamical system, develop custom optimizers and samplers with better properties and apply them to non-smooth, non-convex problems from an optimization perspective and multi-modal distributions from a sampling perspective.
\end{itemize}
We provide an afformative answer to this quesiton in a modest setting.

\subsection {EM-PSO:}
Exponentially Averaged Momentum Particle Swarm Optimization (EM-PSO) \cite{urvilEMPSO} is a variant of PSO. It adopts PSO’s major advantages, such as higher robustness to local minima. It provides more weight to the exploration part, which is an essential part of optimization problems. It has an additional tunable parameter i.e. exponentially averaged momentum, which adds flexibility to the task of exploration better than PSO or its vanilla momentum version.

The Adam optimizer \cite{kingmaadam} is a combination of SGD with momentum and RMSProp. It leverages the momentum by using the moving average of the gradient instead of the gradient itself like in SGD with momentum, and the squared gradients are used to scale the learning rate like in RMSProp. It calculates adaptive learning rates for different parameters from the estimates of the second and first moments of the gradients. The gradient of any function, differentiable or not, can be approximated by using the parameters of EM-PSO. We can leverage computed, approximated gradients from EM-PSO in the update rule for the Adam optimizer scheme. Experimental results show that this leads to lower execution time and comparable (and most times superior) performance to Adam \cite{adaswarm}.

\subsection{Hamiltonian Monte Carlo (HMC)}
Hamiltonian Monte Carlo (HMC) is an algorithm belonging to a class of algorithms known as Markov Chain Monte Carlo (MCMC) \cite{hmc}. The original MCMC algorithm was devised in $1953$ by Metropolis et al. to simulate the distribution of states for a system of idealized molecules \cite{metropoliscomputing}. In a $1987$ paper by Duane et al., the stochastic MCMC algorithm was combined with the deterministic Hamiltonian algorithm to obtain Hybrid Monte Carlo, presently known as Hamiltonian Monte Carlo \cite{duanehybrid}.

The goal of HMC, and indeed MCMC methods in general, is to sample from a given distribution in such a way that the distribution of the samples is equivalent to the original samples in the limit of samples drawn. The quality of an MCMC algorithm is derived from the speed at which the distribution of samples converges to the a posterior distribution. HMC involves using Hamiltonian dynamics to produce more independent and distant proposals than the vanilla Metropolis algorithm with random walks \cite{hmc}. A requirement of Hamiltonian dynamics, is that along with the position variable, there must be a momentum variable that stands for the momentum of the particle in the real world. However, since the momentum has no real analogue in a Monte Carlo simulation, it is simply considered a fictitious variable and is sampled from a random normal distribution.

\subsection{Extending EM-PSO with HMC}
\label{sec:extension}
One challenge with current MCMC samplers is that they don't scale well for large problems (in data and model size) - modern deep learning architectures being a case in point. HMC, based on Hamiltonian dynamics, is a very efficient MCMC sampler that exploits gradients. A Hamiltonian dynamical systems operates with a d-dimensional \textit{position} vector $q$, and a d-dimensional \textit{momentum} vector $p$ \cite{nealhamiltonian}. This system is described by a the \textit{Hamiltonian} $H(q, p)$. The partial derivatives of $H(q, p)$ determine how $q$ and $p$ change over time $t$, according to Hamilton's equations: $\frac{dq_i}{dt} = \frac{\delta H}{\delta p_i}, \ \frac{dp_i}{dt} = - \frac{\delta H}{\delta q_i}$ for $i = 1,2,3, ... ,d$. We are seeking solutions to the above PDEs. But what is $H(p,q)$ here? For HMC, we typically write the Hamiltonian as: $H(q, p) = U(q) + K(p)$ where, $U(q)$ is called the \textit{potential energy}, and can be defined to be negative the log probability density that we wish to sample from, plus any constant that is convenient. $K(p)$ is called the \textit{kinetic energy}, and is usually defined as $K(p) = \frac{p^T M^{-1} p} {2}$. Here, $M$ is a symmetric, positive-definite "mass matrix", which is typically diagonal, and is often a scalar multiple of the identity matrix. With these forms for $H$ and $K$, Hamilton's equations of motion can be solved, which produce either the sampling paths or the search paths. Typically, the potential energy is the objective function being minimized in the optimization problem. The kinetic energy is somewhat an artificial construct and can be designed for convenience. It is exactly the kinetic energy we can play with.  It may be plausible to couple the dynamics of HMC and PSO. Specifically,  consider the PSO and say, we are tracking a particle, along the $i^{th}$ dimension: $p_i(t + 1) = p_i(t) + c_1 r_1(q^{best} - q(t))_i + c_2 r_2 (g^{best} - q(t))_i \ \& \ q_i(t + 1) =
q_i(t) + p_i(t+1)$. We can now construct the kinetic energy terms in the Hamiltonian, defined in terms of $q$. This is one way we can couple PSO with HMC. However, despite its appeal, in a DNN setting, the parameters can run into billions, and this variant of HMC-PSO may not be practical. Instead, we can consider a hybrid particle swarm, where some particles follow standard EM-PSO dynamics, and another set of particles follow HMC dynamics. The gradients of the potential function of the HMC can now be replaced with approximate gradients coming from the other particle set. We explore the latter approach in this paper.

\section{Our Contribution: Hamiltonian Monte Carlo Particle Swarm Optimizer (HMC-PSO)}
We have considered a few methods of coupling Hamiltonian Monte Carlo with PSO. The first question to ask was whether we would create an entirely new form of generic dynamics that used a ''dial" to decide whether to optimize or sample, i.e., exploit or explore, or something in between? We have not used such a method since we wanted to stick with dynamics that had already been investigated, but we propose this as an avenue to be explored in the future. The methods we devised involved using particles that either used HMC dynamics or PSO dynamics.
\subsection{HMC-PSO : Coupling HMC and PSO}
An important challenge while integrating HMC with EM-PSO was to implement HMC dynamics without using explicitly computed gradients since we would otherwise lose the property of a gradient-free optimizer. As discussed earlier, such coupling may not scale to modern DNNs. The solution that we used was to use a modification of the approximation from EM-PSO for approximating the gradient in place of $U_{grad}$ in the normal HMC. The approximation is given by: 
\begin{align} \label{hmcapprox}
U_{grad}(x) = -\frac{c_1r_1 (p_{best} - x) + c_2r_2 (g_{best} - x)}{\eta}
\end{align}
where $c_1r_1$ and $c_2r_2$ have the same meaning they do in EM-PSO, $\eta$ is the total path length of a leapfrog step in HMC, $g_{best}$ is the best point found by the swarm, and $p_{best}$ is the best point found by the particle. While this would probably not be viable when requiring the samples to conform to a predefined distribution, we only need qualitative similarity between the sample distribution and the loss function. Thus, it appears reasonable to replace momentum and position updates in EM-PSO with HMC dynamics but leave the rest of the PSO machinery intact. The goal is to get a significantly better exploration of the space compared to EM-PSO, including a guarantee that the swarm would continue to explore as long as it runs. However, this happens at the cost of the local optimization capabilities of EM-PSO and speed of convergence (in comparison with EM-PSO). Slower convergence is expected since a single position-velocity-momentum update from EM-PSO is replaced by a significantly longer leapfrog update from HMC-PSO. As a remedy to over-correcting for EM-PSO's poor exploration capabilities, we reinstated EM-PSO while keeping HMC as a vital component. As a result, we devised a scheme to couple a single HMC particle with an EM-PSO swarm. Both these components are tightly coupled by sharing information about the best position found so far. Therefore, if the HMC particle, through its sampling dynamics, finds a better spot even after the swarm has settled at a local minimum, the swarm will migrate to this new minimum. In turn, the information about this $g_{best}$ position will be used to guide the HMC particle via the EM-PSO approximation. The detailed HMC-PSO algorithm is given in Algorithm $1$.

\begin{algorithm}
\caption{HMC-PSO}
\SetKwInOut{Input}{input}
\SetKwInOut{Output}{output}
\Input{Number of particles in the swarm, N}
\Input{Fitness function, fitness}
\Input{Parameters of PSO, c1 and c2}
$g_{best}.value$ = $\infty$
$g_{best}.position$ = undefined
swarm $\leftarrow$ [N EM particles]
hmc $\leftarrow$ HMC particle
\While{$g_{best}$ \emph{has not converged}}
{
    \For{\emph{Particle} $p_i$ \emph{in swarm, including hmc}}{
        $f$ = fitness($p_i$)\\
        \If{$f < p_i$.$p_{best}.value$}{
            $p_i$.$p_{best}.value$ = $f$\\
            $p_i$.$p_{best}.position$ = $p_i.position$\\
            \If{$f < g_{best}.value$}{
                $g_{best}.value$ = $f$\\
                $g_{best}.position$ = $p_i.position$\\
            }
        }
    }
    \For{\emph{Particle} $p_i$ \emph{in swarm}}{
        $p_i$.move($g_{best}$)      \tcp*{move $p_i$ using EM-PSO approximation with $g_{best}$}
    }
hmc.move($g_{best}$)}
\end{algorithm}

It might be noted that the EM-PSO approximation of the gradient leads to a loss of information for the HMC particle. Whereas earlier, the particle would be informed by the local landscape of the potential, it now receives information solely through $g_{best}$. As a result, upon convergence of the EM-PSO swarm, the HMC particle will behave as if it has been trapped in a potential centered around $g_{best}$. We can investigate the form of this pseudo-potential. The gradient update is given by Equation \ref{hmcapprox}. We can see that if the HMC particle comes sufficiently close to $g_{best}$ since the gradient is directed towards it, we will have $p_{best} \approx g_{best}$. As a result, the approximation will have the form:
\begin{align}
    U_{grad}(x) &= K(x - x_0) \\
\textrm{where, } x_0 &= g_{best} \\
K &= \frac{c_1r_1 + c_2r_2}{\eta}
\end{align}

Thus, the HMC particle may become trapped in a sort of randomized harmonic oscillator centered at $g_{best}$. However, this results in an exploration centered around the best known optimum. From elementary physics, we know that the resulting probability distribution for the HMC particle will be a gaussian centered at $g_{best}$. In fact, under such conditions, the draws from HMC can be used to perform a Laplce Approximation of the posteior density around a mode \cite{laplace}.

\subsection{Complexity Analysis}
From Algorithm $1$, we see that the initialization steps $1$ and $2$ are $O(1)$ whereas the step $3$ is $O(N)$, where $N$ is the number of particles in the swarm. Similarly, step $4$ i.e. initializing $N$ EM particles takes $O(N)$ time \cite{urvilEMPSO}. Step $5$ is $O(1)$. With regards to the while loop, let us assume that it takes $t$ iterations to complete. We have two for loops on step $6$ and step $17$; they both take $O(N)$ time to complete, since all operations inside these loops are $O(1)$. Step $20$ takes $O(L)$ time, where $L$ is the number of steps per sample for Hamiltonian Monte Carlo.
Hence, the overall complexity is $O(T(L+N))$. 

\subsection{HMC-PSO solver for Golomb's Ruler}

A Golomb ruler is a set of distinct non-negative integers where the difference between every pair of integers is unique. This can be alternatively described by an imaginary ruler with integral markings such that no two pairs of marks are apart by the same distance. The order of the ruler is defined to be the number of markings, and the length as the largest marking. A Golomb Ruler is said to be \textit{optimal} if no ruler of shorter length exists for the given order. An intuitive way of solving this problem would be by using powers of $2$ as the markings. While this ensures a solution, it's length would be too large to be optimal. Furthermore, the computational complexity of the problem increases exponentially as the order of the ruler increases. The Traditional search-based algorithms like backtracking and branch-and-bound would not scale due to this computational complexity. Golomb's ruler finds application in several areas including error-correcting codes,  selecting radio frequencies to reduce the effects of intermodulation interference with both terrestrial and extraterrestrial applications, design of phased arrays of radio antennas etc. Multi-ratio current transformers use Golomb's rulers to place transformer tap points.

The Golomb's Ruler problem is formulated by using a unconstrained optimization problem \cite{prudhvidqn}, which we solve by the HMC-PSO algorithm. Given the Golomb Ruler $G$, we measure the number of violations caused to the rule, i.e. a common differences obtained within markings.

Let $V_G(d)$ be the violation score for a distance $d$. It is the number of times the distance $d$ appears between any two markings of the ruler $G$. $V_G(d)$ is defined as:
\begin{align}
V_G(d) = \max(0, | d_{ij} = d \ : \ 1 \leq j < i \leq m| - 1)
\end{align}The total violation of the ruler $V(G)$ can thus be given as:
\begin{align}
V(G) = \sum_{d=1}^n V_G(d)
\end{align}for a sufficiently large $n$. As closer we are to $0$, closer we are to a Golomb Ruler. If $V(G) = 0$, it means we have successfully found a Golomb Ruler for the given order. For our experiment, we aim to achieve two important objectives: \\1. For the given order $n$, find a Golomb Ruler (i.e. a ruler with $V(G) = 0$) \& \\2. Find the optimal or close to optimal Golomb ruler (i.e. a ruler with $V(G) = 0$ and smallest length).
 
It's not possible to use the total violation as the cost function for the optimization procedure directly as $V(G)$ can't differentiate between a valid ruler of smaller length and a valid ruler of larger length. We add a term proportional to the length of the proposed ruler.
Let $n$ be the order of $G$, the cost function for Golomb Ruler optimization i.e. $L(G)$ is defined as:
\begin{align}
L(G) = V(G) + 10^{-k}\max(G)
\end{align}where, $k$ is the smallest positive integer such that $0 < 10^{-k} \max(G) <  1$. Intuitively, we pick a $k$ such that the largest element obtained in the Golomb's Ruler in any iteration (while running HMC-PSO) when multiplied with $10^{-k}$ does not exceed $1$. We do this to ensure priority of validness over length of $G$. Note that, $V(G)$ takes positive integer and $0$ values. Thus $L(G)$ ensures that the loss via violation always remains greater than loss due to the ruler being large. This is essential since we need a Golomb's ruler rather than a small ruler with $1$ violation. For example, if in the entire optimization, maximum length for the ruler reached is smaller than $1000$, we can safely take $k = 3$, since we know that for the order of rulers lesser than $11$, the optimal ruler has a length of $72$ and hence, we will likely get rulers of length lesser than $1000$. Taking $k=3$ would ensure that $0 < 10^{-3} \max(G) <  1$ and thus $L(G)$ always penalizes a violation more than the length of the ruler.

\noindent \textbf{Optimization Procedure}: The space over which optimization is done is $\mathbb{R}^n$. Any position vector is given by $\boldsymbol{p} = (p_1, p_2, p_3, ... , p_n)$. The Golomb ruler corresponding to this position vector is given by:
\begin{equation*}
\boldsymbol{p} = (p_1, p_2, p_3, ... , p_n) <=> G =  \lfloor \ || (p_1, p_2, p_3, ... , p_n) || \ \rfloor
\end{equation*}where, $||.||$ is the absolute value function and $\lfloor . \rfloor$ is the floor function. A set of particles start at some initial position and follow the HMC-PSO rule to get to better positions based on $L(G)$.

\subsection{HMC-PSO solver with DNNs for Classification tasks}


In the previous section, we aimed to utilize HMC-PSO as a direct optimizer for solving the optimization problem. However, we can also use HMC-PSO to train DNNs. There is an important advantage to doing this\; PSO techniques do not rely on gradients. This means it is possible to use PSO on landscapes that do not rend well to gradient computation. Another advantage, something that has been found empirically, is that PSO allows the network to make better steps with the data given, so as to improve the loss faster than traditional methods of training. We shall elaborate on a method in which we can train DNNs using HMC-PSO.

We use the labels of the data to initialize a swarm of particles in such a way that they are close to a minimum of the loss function. The particles then follow HMC-PSO dynamics in order to find the best possible location in the vicinity of the initialization, $g_{best}$. The information about $g_{best}$ is then used to make a first order estimation of the gradient of the loss function and make a step towards $g_{best}$. This is done because of an equivalence between EM-PSO and Stochastic Gradient Descent \cite{adaswarm}. The estimate is given by:
\begin{align}
    grad = -\frac{(c_1r_1 + c_2r_2)}{\epsilon} (g_{best} - y)
\end{align}where, $c_1r_1$ and $c_2r_2$ are the values obtained in the last iteration of Algorithm $1$ and $y$ is the value output by the last layer after forward propagation. This estimate is then used to update the weights of the rest of the network through backpropagation by the Adam optimizer. It has been observed that this technique provides improved performance over using Adam with numerically computed gradients \cite{adaswarm}.

It is important to note that the initialization of the swarm is an essential step since EM-PSO heavily relies on the initialization to produce a decent $g_{best}$ since it almost always settles into the nearest local minimum. Each particle in the swarm has dimensions $batch\_size \times num\_classes$, where $batch\_size$ is the number of data items in a mini-batch, and $num\_classes$ is the number of neurons in the last layer. Each of the $batch\_size$ rows is initialized according to the index of the label. The values of the position matrix are set to 1 at the target index and -4 in all other columns of the row. Intuitively, we can understand how this incentivizes the network to learn the correct mapping from data to labels.

In this paper, we have utilized a Resnet-18 architecture for classification task on image-based datasets like: MNIST, Fashion MNIST and CIFAR-10. The results are mentioned in the next section.

\section{Results}
\label{sec:results}

Before reporting the results of HMC-PSO, we note that: \textit{1. EM-PSO couldn't solve the Golomb ruler problem 2. EM-PSO is just an approximation of first order derivative and was not exploited in the literature to solve Deep NN based classification problems}.

\subsection{Golomb Ruler}

P. Prudhvi et al. \cite{prudhvidqn} utilized the use of Deep Q-Learning Networks (DQN) to solve the Golomb's Ruler problem. It used $-V(G)$ as the reward score. The agent repeatedly takes actions until all the markings are predicted and the violation score $V(G)$ is $0$. This would mark the end of $1$ episode and the agent is trained over a total of $100$ episodes. The replay phase trains the agent to better predict the
markings of a ruler.

A comparison of the DQN approach with standard solution approaches to solve constraint satisfaction problems, such as backtracking and branch-and-bound, demonstrated the efficacy of the DQN approach, with significant computational savings as the order of the problem increases. The results of this experiment by \cite{prudhvidqn} (for the $22^{nd}$ episode) are provided below:

\begin{center}
\begin{tabular}{||c | c | c | c||} 
 \hline
 N-Value & Backtracking & Branch and Bound & DQN\\ [0.5ex] 
 \hline\hline
 4 & 6 & 6 & 4 \\ 
 \hline
 5 & 3615 & 13 & 5\\
 \hline
 6 & 170264 & 48 & 6 \\
 \hline
 7 & 10095777 & 268 & 7 \\
 \hline
\end{tabular}
\end{center}

The above table showcases the number of iterations required to converge to a solution by various algorithms: DQN shows a significant improvement compared to other traditional algorithms like Backtracking and Branch and Bound. However, there is a major limitation for this technique. All methods analysed (including DQN) only arrive at a valid solution for the Golomb's ruler problem. It does not find an optimal Golomb's Ruler or rulers close to the optimal ruler.

Running HMC-PSO, we observe that uptil order $n = 50$, the HMC-PSO obtains a loss $L(G) < 1$ i.e. a valid Golomb Ruler within the first iteration itself. Hence, HMC-PSO achieves the same objectives that DQN and other algorithms aim in just $1$ iteration and for a significantly higher order rulers. This shows that HMC-PSO is able to scale well on problems which require unimaginable number of computations (it's very difficult to estimate the number of iterations a backtracking algorithm would require for a Golomb ruler of order $50$ if it takes $10095777$ iterations for a ruler of order $7$). Further, HMC-PSO then focuses on finding the optimal ruler, and is able to find rulers close to the optimal ruler. The number of iterations taken vs the loss obtained for Golomb ruler for different values of order $n$ is given as follows in Table 1.
\begin{table}
\caption{HMC-PSO's results on the Golomb Ruler problem}
\centering
\begin{tabular}{||c | c | c | c ||} 
 \hline
 Order & Iterations & Loss obtained & Optimal loss\\ [0.5ex] 
 \hline\hline
 5 & 3 & 0.011 & 0.011 \\ 
 \hline
 7 & 65 & 0.025 & 0.025\\
 \hline
 9 & 80 & 0.060 & 0.044 \\
 \hline
 11 & 80 & 0.106 & 0.072 \\
 \hline
 15 & 80 & 0.395 & 0.151 \\
 \hline
\end{tabular}
\end{table}
For these results, we notice that the length of Golomb Ruler never crosses $1000$, and hence we take $k=3$. We notice that HMC-PSO is able to find optimal Golomb ruler of order $7$ and close to optimal but valid rulers for higher orders. The decimal digits of loss is nothing but the length of the ruler. For instance, for $n=11$, a loss of $0.106$ signifies that a valid Golomb ruler of length $106$ is found after running $80$ iterations. The optimal loss of $0.072$ means that the optimal ruler for order $n=11$ has a length of $72$.

\subsection{Classification problems}

We tested the optimizers HMC-PSO, Adam, AdaGrad, AdaDelta, RMSProp, and NAdam in a DNN setting i.e. ResNet 18 architecture on the MNIST Digit Recognition, Fashion MNIST and CIFAR10 datasets. We trained the model for $20$ epochs for each of the optimizers. The results are briefly mentioned in Table $2$ and $3$. The performance of HMC-PSO is better than or equivalent to other state-of-the-art optimizers.
\begin{table}
\caption{Testing accuracy comparison of the optimizers based on Cross-Entropy Loss.}
\centering
\begin{tabular}{||c | c | c | c | c | c | c ||} 
 \hline
 Dataset & HMC-PSO & Adam & RMSProp & NAdam & AdaGrad & AdaDelta\\ [0.5ex] 
 \hline\hline
 MNIST Digit & \textbf{99.25\%} & 98.63\%  & 98.81\% & 98.85\% & 99.2\% & 97.24\% \\ 
 \hline
 Fashion MNIST & 93.62\% & 93.40\% & \textbf{93.88\%} & 93.17\% & 92.89\% & 85.79\%\\
 \hline
 CIFAR-10 & \textbf{89.04\%} & 87.67\% & 87.90\% & 87.98\% & 80.39\% & 56.59\%\\
 \hline
\end{tabular}
\end{table}

\begin{table}
\caption{Testing accuracy comparison of the optimizers based on Multi-Margin Loss.}
\centering
\begin{tabular}{||c | c | c | c | c | c | c ||} 
 \hline
 Dataset & HMC-PSO & Adam & RMSProp & NAdam & AdaGrad & AdaDelta\\ [0.5ex] 
 \hline\hline
 MNIST Digit & 98.19\% & 97.76\%  & 98.24\% & 97.75\% & 98.54\% & \textbf{98.88}\% \\ 
 \hline
 Fashion MNIST & \textbf{93.43\%} & 88.53\% & 91.13\% & 91.17\% & 91.15\% & 93.31\%\\
 \hline
 CIFAR-10 & \textbf{82.33\%} & 78.19\% & 76.09\% & 80.26\% & 76.76\% & 78.24\%\\
 \hline
\end{tabular}
\end{table}
\section{Conclusion}
\label{sec:conclusion}

In summary, we propose Hamiltonian Monte Carlo Particle Swarm Optimization (HMC-PSO) - an optimization algorithm that can either be directly used for any optimization procedure or can be utilized as a tool to approximate gradients of any differentiable or non-differentiable loss function and thus used alongside DNNs just like other SOTA optimizers. Because of coupling btween HMC and PSO, the HMC-PSO algorithm reaps considerable benefits from both: extensively searching the parameter space and converging quickly to the minima. Apart from this, HMC-PSO acquires an essential benefit from EM-PSO: it is able to approximate gradients for non-continuous and non-differentiable loss functions which usual gradient-descent algorithms are unable to compute. This is one of the reasons why HMC-PSO stands out from other SOTA optimizers apart from better results.

In this paper, we focused on the prior-work i.e. EM-PSO as the motivation for our research. We provided necessary motivation as to why we utilized HMC and how we successfully coupled HMC and PSO techniques. Further, we discussed two different uses of HMC-PSO: as a direct optimization procedure for solving the Golomb's Ruler problem and as a optimizer for DNN in  classification problems. We obtained better results than other SOTA methods.

An essential point of difference between EM-PSO and HMC-PSO is that HMC-PSO explores the search-space much more extensively and ensures in almost every scenario that the global optimum is found. This was observed when we tested EM-PSO and HMC-PSO on $9$ different multi-modal Gaussian distributions. EM-PSO, due to it's first-order convergence nature, was only able to converge linearly to the local maxima, whereas HMC-PSO effectively explored and found all the maxima and converged to the global maxima. HMC-PSO produced effective test-results on 1-d benchmark optimization problems as well (Results are stored in a Github Link:.\cite{githublink}).


%
%
%
%

\end{document}